\documentclass[lettersize,journal]{IEEEtran}
\usepackage{amsmath,amsfonts}
\usepackage{algorithmic}
\usepackage{algorithm}
\usepackage{array}
\usepackage[caption=false,font=normalsize,labelfont=sf,textfont=sf]{subfig}
\usepackage{textcomp}
\usepackage{stfloats}
\usepackage{url}
\usepackage{verbatim}
\usepackage{graphicx}
\usepackage{cite}
\usepackage{xcolor}

\usepackage{arydshln}
\usepackage{multirow}
\usepackage{booktabs}
\begin{document}

\title{Unified Representation Space for 3D Visual Grounding}

\author{Yinuo Zheng, Lipeng Gu, Honghua Chen, Liangliang Nan, and Mingqiang Wei,~\textit{Senior Member, IEEE}
\thanks{Y. Zheng, L. Gu, H. Chen and M. Wei are with the School
of Computer Science and Technology, Nanjing University of Aeronautics
and Astronautics, Nanjing, China (e-mail: zhyinuo@nuaa.edu.cn; glp1224@163.com; chenhonghuacn@gmail.com; mingqiang.wei@gmail.com).}
\thanks{L. Nan is with the Urban Data Science Section, Delft University of Technology, Delft, Netherlands (e-mail: liangliang.nan@tudelft.nl).}
}

\markboth{Journal of \LaTeX\ Class Files,~Vol.~14, No.~8, August~2021}%
{Shell \MakeLowercase{\textit{et al.}}: A Sample Article Using IEEEtran.cls for IEEE Journals}

\IEEEpubid{0000--0000/00\$00.00~\copyright~2021 IEEE}

\maketitle

\begin{abstract}
    3D visual grounding (3DVG) is a critical task in scene understanding that aims to identify objects in 3D scenes based on text descriptions.
    However, existing methods rely on separately pre-trained vision and text encoders, resulting in a significant gap between the two modalities in terms of spatial geometry and semantic categories. This discrepancy often causes errors in object positioning and classification.
    The paper proposes \textbf{UniSpace-3D}, which innovatively introduces a \textbf{uni}fied representation \textbf{space} for \textbf{3D}VG, effectively bridging the gap between visual and textual features.
    Specifically, UniSpace-3D incorporates three innovative designs: 
    i) a unified representation encoder that leverages the pre-trained CLIP model to map visual and textual features into a unified representation space, effectively bridging the gap between the two modalities;
    ii) a multi-modal contrastive learning module that further reduces the modality gap;
    iii) a language-guided query selection module that utilizes the positional and semantic information to identify object candidate points aligned with textual descriptions. 
    Extensive experiments demonstrate that UniSpace-3D outperforms baseline models by at least 2.24\% on the ScanRefer and Nr3D/Sr3D datasets. The code will be made available upon acceptance of the paper.

\end{abstract}


\section{Introduction} 
\IEEEPARstart{P}oint clouds have become a foundational 3D geometric data representation in computer graphics and computer vision, with applications in various domains, including archaeology~\cite{DBLP:archaeology}, augmented reality~\cite{realityaug}, autonomous driving~\cite{vote},  robotic navigation~\cite{heterogeneous, continuous}. Building on this, 3D visual grounding, which aims to localize an object in 3D scenes based on a textual description, has become a crucial challenge at the intersection of language and spatial reasoning\cite{human,tree}. It emphasizes the interaction between language and spatial understanding.

\par Recent advancement in 3DVG can be categorized into two-stage and one-stage architectures. 
Early works like ScanRefer~\cite{scanrefer}, 3DVGTrans~\cite{3dvg}, and SAT~\cite{sat} adopt the two-stage architecture, first using pre-trained object detectors to generate candidate bounding boxes and then selecting the object from these candidates. 
Given that the performance of two-stage methods is heavily dependent on the quality of the detectors, one-stage methods, such as 3DSPS~\cite{3dsps}, which directly localize objects through language-guided keypoint detection, have gained increasing attention. Despite these advancements, a critical issue persists in 3DVG: \textit{inaccurate localization alongside correct classification.}
\begin{figure}[tbp]
  \centering
  \includegraphics[width=.95\linewidth]{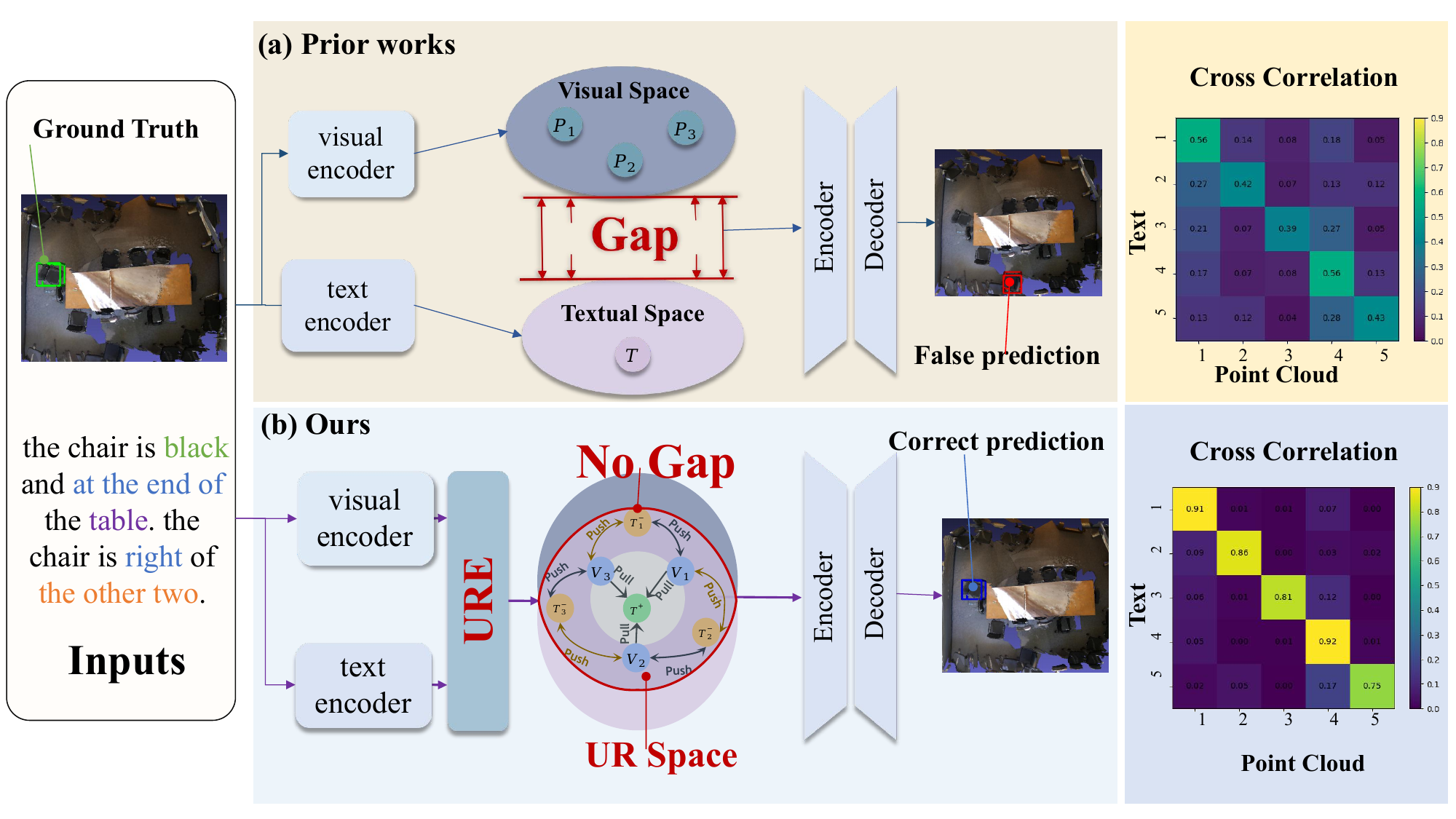}
  \caption{ \textbf{Comparison between the prior works (a) and ours (b).} 
  Our method achieves more accurate grounding by bridging the gap between the visual and textual features into the unified representation (UR) space.
  As illustrated on the right, mapping text and point clouds into the UR space enhances cross-model correlation, with yellow indicating stronger alignment.
}
\label{fig:intro}
\end{figure}

\par There are two main reasons for the above challenge.
First, existing methods, such as EDA\cite{vpp} and VPP-Net\cite{eda}, rely on separately pre-trained visual and textual encoders that independently capture positional and semantic features from point clouds and text, respectively. This leads to a significant gap between the two modalities (as shown in Fig.~\textcolor{red}{\ref{fig:intro}}).
Second, text descriptions and point clouds both contain positional and semantic information about the 3D scene. However, existing methods rely solely on visual queries to select object candidate points and fail to fully leverage the positional and semantic information embedded in the text modality. This limitation negatively impacts object localization performance.

\IEEEpubidadjcol

\par We propose UniSpace-3D, a unified representation space for 3DVG.
UniSpace-3D bridges the gap between visual and textual feature spaces by aligning positional and semantic information from both modalities. 
our method is built on three key components: the unified representation encoder (URE), the multi-modal contrastive learning (MMCL) module, and the language-guided query selection (LGQS) module.
Specifically, the URE maps the positional and semantic information from point clouds and text into a unified representation (UR) space. 
The MMCL further reduces the disparity between visual and textual features in the UR space by enhancing consistency. It achieves this by bringing visual embedding closer to their corresponding textual embeddings while pushing them away from unrelated textual embeddings.
Finally, LGQS utilizes the positional and semantic information from both modalities to accurately identify object candidate points that align with the text description. This step reduces localization errors and improves grounding accuracy. Extensive experiments show that UniSpace-3D outperforms baseline models by at least 2.24\% on the ScanRefer and Nr3D/Sr3D datasets. 

Our contributions can be summarized as:
\begin{itemize}
    \item We propose the URE module, which maps visual and textual features into a unified representation space, effectively bridging the gap between modalities;
    \item We introduce the MMCL module, which further reduces the gap between visual and textual representations, enabling effective positional and semantic alignment between both modalities;
    \item We propose the LGQS module, which improves object localization by focusing on object candidate points that match the positional and semantic information in the text, ensuring the precise identification and localization of the object described in the text.
\end{itemize}
\par The remainder of this paper is organized as follows: Section \ref{sec:relatedWork} introduces the related work. Section \ref{sec:Method}  gives the details of our method. Section \ref{sec:Experiments} shows the experiments of our method, followed by conclusion in Section \ref{sec:Conculsion}. 

\IEEEpubidadjcol

\section{Related Work} \label{sec:relatedWork}
\subsection{3D Vision-Language Tasks}
\par Vision and language are the two most fundamental modalities to understand and interact with the 3D real world, giving rise to a variety of 3D vision-language tasks.
3D dense captioning \cite{Scan2Cap, X-Trans2Cap, Vote2Cap} involves identifying all objects in complex 3D scenes and generating descriptive captions. 3D visual grounding \cite{scanrefer, Referit3d, bottom} takes 3D point clouds and language descriptions to localize the target objects via bounding boxes. 
3D question answering \cite{ScanQA, SQA3d} addresses answering questions based on visual information from 3D scenes.
All these tasks primarily focus on aligning visual and linguistic features, particularly spatial and semantic information.
In this work, we focus on the fundamental task of 3D visual grounding (3DVG), enabling machines to comprehend both 3D point clouds and natural language simultaneously.

\subsection{3D Visual Grounding}
\par 3D visual grounding aims to localize the corresponding 3D proposal described by the input sentence. In contrast to 2D images, point clouds exhibit characteristics of sparsity and noise, lacking dense texture and structured representation. These attributes seriously limit the migration of advanced 2D localization methods, which rely on pixel-level visual encoding. 
The main datasets for 3DVG include ReferIt3d \cite{Referit3d} and ScanRefer \cite{scanrefer}. These datasets are derived from ScanNet \cite{scannet}.
According to the overall model architecture,  previous works can be divided into two distinct groups: two-stage methods and one-stage methods.

\par \textbf{Two-Stage Methods}~ Most existing 3DVG methods adopt a two-stage framework \cite{scanrefer, 3dvg, sat, multiviewtrans}. Scanrefer~\cite{scanrefer} first utilizes a 3D object detector to generate object proposals and subsequently identifies the target proposal that corresponds to the given query. 
SAT~\cite{sat} leverages 2D semantics to assist 3D representation learning.
SeCG~\cite{secg} propose a graph-based model to enhance cross-modal alignment 
.
Some recent works \cite{3dvg, 3drpnet, vpp}
utilize transformers\cite{transformer} as a key module to accomplish the modality alignment.
However, the performance of these models depends heavily on the quality of the proposals produced in the first stage. In order to solve this problem, Single-stage methods are introduced.

\par \textbf{Single-Stage Methods} 
Without relying on the quality of pre-trained object generators (i.e., 3D detectors or segmentors), recent 3D visual grounding methods follow a one-stage framework that trains grounding models end-to-end, from feature extraction to final cross-modal grounding.
Compared to previous detection-based frameworks, this model is more efficient as it eliminates the need for complex reasoning across multiple object proposals. 
3D-SPS~\cite{3dsps} proposed a one-stage method that directly infers the locations of objects from the point cloud.
BUTD-DETR~\cite{bottom} encodes the box proposal tokens and decodes objects from contextualized features. 
Following 3D-SPS~\cite{3dsps}, to better align visual language features, EDA~\cite{eda} proposes a text decoupling module to parse language descriptions into multiple semantic components.

These methods show impressive results. However, aligning features from different modalities remains challenging due to the inevitable feature gap between textual and visual spatial-semantic information.  
To address this, we propose a unified representation space for 3DVG to effectively integrate separate feature spaces and identify object candidate points aligned with the input text, enabling accurate grounding.

\begin{figure*}[htb]
  \centering
  \includegraphics[width=.95\linewidth]{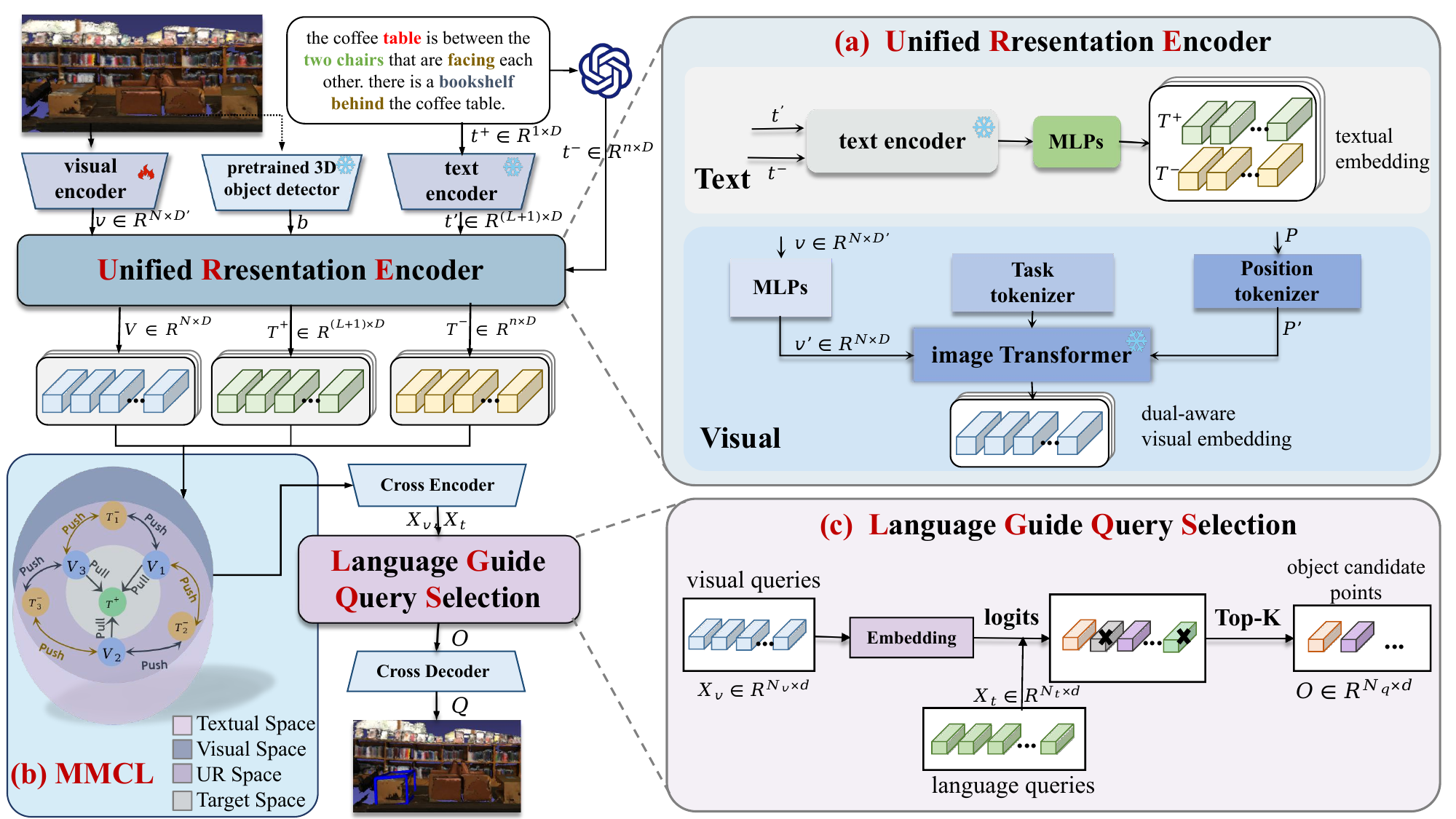}
  \caption{\label{fig:main}
        \textbf{Overview of UniSpace-3D.} The framework comprises three key components. (a) URE, which maps features from point clouds and text descriptions into a unified representation space. (b) MMCL, which refines alignment by reducing the gap between visual and textual embeddings. (c) LGQS, which identifies object candidate points that closely correspond to the text descriptions, enhancing grounding accuracy.
}
\end{figure*}

\section{Proposed Method} \label{sec:Method}
\textbf{Overview.} Existing 3DVG methods rely on independently pre-trained feature encoders to capture positional and semantic information, resulting in a considerable gap between the two modalities.
As shown in Fig.~\textcolor{red}{\ref{fig:intro}}, this gap is the key factor causing correct classification but inaccurate localization in 3DVG, a challenge that many existing methods fail to address.
To overcome this challenge, we propose UniSpace-3D.

As shown in Fig.~\textcolor{red}{\ref{fig:main}}, UniSpace-3D incorporates three innovative designs. 
First, the unified representation encoder (URE, see Sec. \textcolor{red}{\ref{sec32}})  effectively captures task- and position-aware visual and textual embeddings within a unified representation (UR) space.
Second, the multi-modal contrastive learning module (MMCL, see Sec. \textcolor{red}{\ref{sec33}}) reduces the remaining feature gap by pulling visual embeddings closer to their corresponding textual embeddings while pushing them away from unrelated textual embeddings.
Finally, the language-guided query selection module (LGQS, see Sec. \textcolor{red}{\ref{sec34}}) selects object candidate points that better align with the text description, enhancing grounding accuracy. We explained the design of our loss function in Sec. \textcolor{red}{\ref{sec35}}. 
Through these innovations, our UniSpace-3D achieves more accurate grounding.

\subsection{Unified Representation Encoder}
\label{sec32}

The quality of extracted visual and textual features significantly impacts 3DVG performance.
However, the disparate spaces of visual and textual features make alignment and understanding challenging. 
To tackle this issue, URE narrows the gap between the disparate feature spaces, thereby enhancing the model's understanding of both the positional and semantic information in each modality. 

Before URE, the input data are first tokenized into text and visual tokens. 
These tokens are fed into the URE to obtain textual embeddings and the task-position dual-aware visual embeddings, both aligned in the same CLIP \cite{clip} space and interpreted as the UR space for 3DVG.

\subsubsection{Tokenization}
The input text and 3D point clouds are encoded by the text encoder and the visual encoder to produce text tokens $t^{\prime}=(t_{cls},t_1,...,t_L)$ and visual tokens $v =(v_1,...,v_N)$. Here, $t_i$ and $v_i$ are the features of each token,  $t_{cls} \in R^D$ is a special token for text classification, and $L$ represents the length of the text description corresponding to the specified target object. In our experiment, the text encoder and the visual encoder are composed of the pre-trained RoBERTa~\cite{roberta} and PointNet++~\cite{pointnet}. In addition, the GroupFree~\cite{groupfree} detector is optionally used to detect a 3D box according to  \cite{eda}, which is subsequently encoded as a box token $b\in R ^{d\times D}$. Here, $d$ is the number of detection boxes and $D$ is the feature dimension.

\subsubsection{Textual Embedding}
The text token is fed directly into the CLIP text encoder to obtain the textual embeddings $T^{+}\in R^{(L+1)\times D} $ and $T^{-}=(T_1^{cls},..., T_n^{cls})$ in UR space. Here, $n $ denotes the number of negative sentences, and $T_i^{cls}\in R^D$ is the token for text classification for each negative sentence, which is detailed in Sec. \textcolor{red}{\ref{sec33}}. 
The textual embeddings consist of positive embeddings \(T^+\) and negative embeddings \(T^-\).

\begin{figure*}[!htb]
  \centering
\includegraphics[width=.95\linewidth]{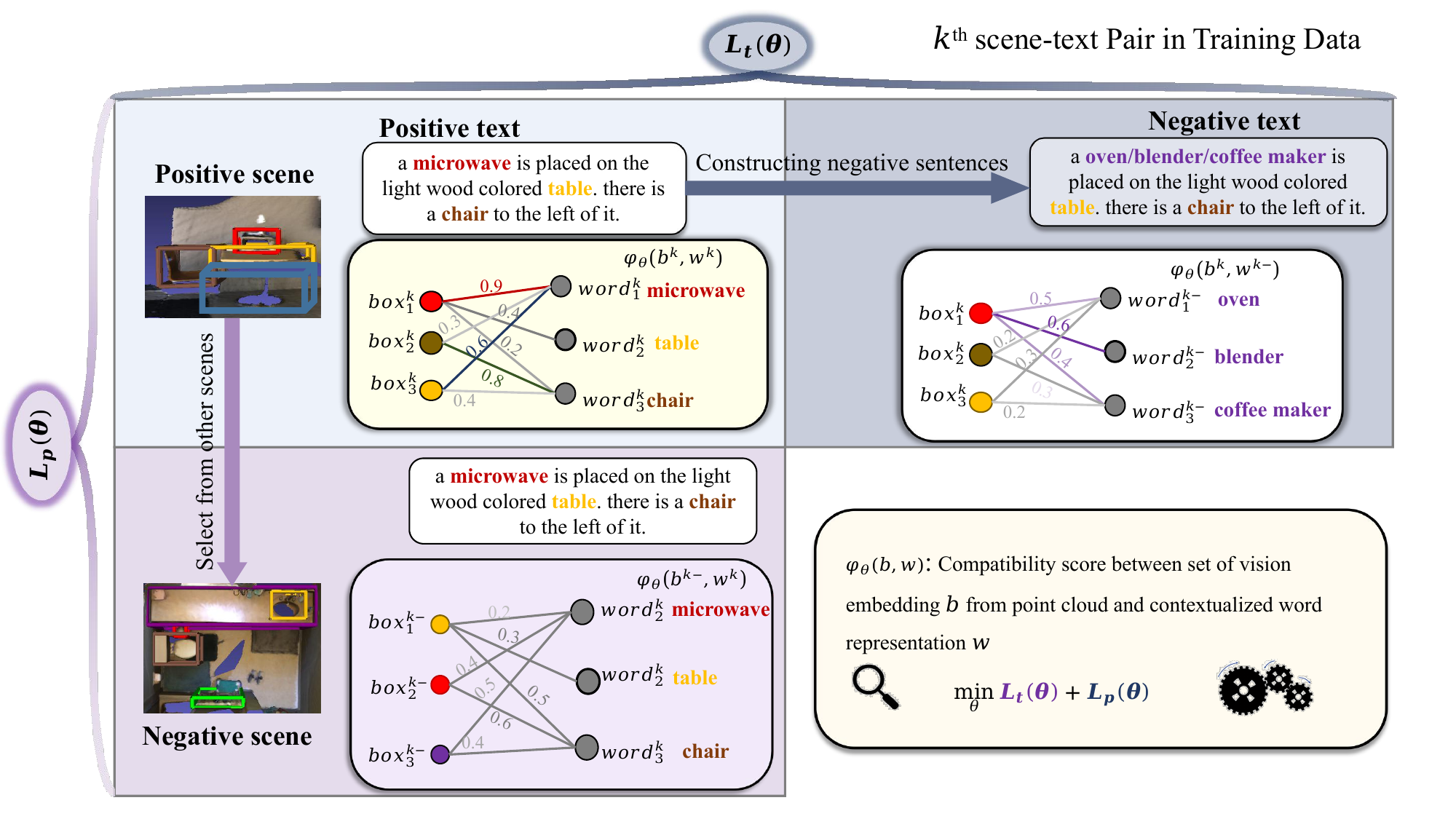}
  \caption{\label{fig:neg1}
    \textbf{Negative contrastive learning in Multi-Modal Contrastive Learning module}.
    MMCL encourages higher compatibility scores between the true grounding scene and the corresponding sentence while discouraging mismatched pairs.
    }
\end{figure*}

\subsubsection{Visual Embedding}

Inspired by EPCL~\cite{epcl}, we use a frozen CLIP model to extract shape-based features from point clouds. CLIP image transformer, trained on image-text pairs, maps tokens $X\in\Omega_I$ to $Y\in\Omega_O$. Similarly, UniSpace-3D leverages PointNet~\cite{pointnet} to map local point cloud patches, viewed as 2D manifolds, into the vision token space $\Omega_I^P$, enabling effective learning.

To align visual tokens into the UR space, we first pass visual tokens $v$ through several MLPs for dimensional transformation, resulting in $v^{\prime} \in \mathbb{R}^{N \times D}$, and then embed $v^{\prime}$ into the CLIP.
However, since CLIP~\cite{clip} is trained on a large dataset of text-image pairs, it lacks specific task information. To address this, we design a task tokenizer to embed point clouds into the UR space for 3DVG tasks. The task tokenizer, implemented as a fully connected layer with learnable parameters, captures global task-related biases. Following \cite{P-Tuning-v2}, we initialize the task token as an enumerator. After transforming the input point cloud into visual tokens $v^{\prime}$, these visual tokens, along with task and position tokens, are fed into the CLIP image transformer to extract task-position dual-aware visual embeddings $V \in \mathbb{R}^{N \times D}$. The transformer is initialized with pre-trained CLIP weights and remains frozen during training.

Fig.~\textcolor{red}{\ref{fig:intro}} shows that the URE can weakly align the text tokens and visual tokens. Before applying URE, the text and visual embedding for the same scene exhibit lower cross-correlation. In contrast, after URE, the text and visual embedding
achieve a higher cross-correlation, indicating improved alignment within the same scene.

\subsection{Multi-Modal Contrastive Learning} 
\label{sec33}
After mapping the visual and text tokens into the UR space, we aim to minimize the remaining feature gap between the two modalities.  To achieve this, we propose the MMCL module that pulls visual embeddings closer to their corresponding textual embeddings while pushing them apart from unrelated textual embeddings.
Specifically, we design the multi-modal contrastive learning loss in Eq. \textcolor{red}{\ref{con_pa}} to achieve this alignment.

\subsubsection{Total Contrastive Loss}

The total contrastive loss is defined as 
\begin{equation}
\mathcal{L}_{cos}= \mathcal{L}_{pos} + \alpha \mathcal{L}_{p} + \beta \mathcal{L}_{t},
\label{con_pa}
\end{equation}
where $\alpha$ and $\beta$ are the weights of different loss rates. The components $\mathcal{L}_{pos}$, $\mathcal{L}_{p}$ and $\mathcal{L}_{t}$ are introduced as follow. 

\subsubsection{Positive Contrastive Loss}
To help learn better multi-modal embeddings, we introduce a positive contrastive loss, defined in Eq. \textcolor{red}{\ref{co:pos}}, to align visual and textual embeddings as 
\begin{equation}
\mathcal{L}_{pos} = \frac{\mathcal{L}_{c}^{T\to V} + \mathcal{L}_{c}^{V\to T}}{2}
\label{co:pos}
\end{equation}
where 
\begin{equation}
\mathcal{L}_{c}^{V\to T} = -\log{\frac{exp(cos(\bar{V_{i} } ,T_{i} )/\tau )}{ {\textstyle \sum_{i=1}^{n}exp(cos(\bar{V_{i} } ,T_{j} )/\tau )} } }
\label{co:lcvt}
\end{equation}
and
\begin{equation}
\mathcal{L}_{c}^{T\to V} = -\log{\frac{exp(cos(T_{i},\bar{V_{i} } )/\tau )}{ {\textstyle \sum_{i=1}^{n}exp(cos(T_{i},\bar{V_{j} })/\tau )} } }
\label{co:lctv}
\end{equation}

Here, $T$ is textual embedding, $\bar{V}$ is the mean of visual embeddings of all target objects paired with a description, and $\tau$ is a temperature parameter.

\subsubsection{Negative Contrastive Loss}

To further reduce the gap between visual and textual embeddings, we leverage contrastive learning~\cite{contrastive} to push visual embeddings apart from unrelated textual embeddings. 
As illustrated in Fig.~\textcolor{red}{\ref{fig:neg1}}, the negative contrastive loss 
consists of two components: $L_{p}$ and the $L_{t}$, details are as follows.

Specifically, the compatibility score $\phi _{\theta } \left ( b,w \right )$ measures the alignment between visual embeddings \(b\) from the scenes and the contextualized word representation \( w \). It is defined as:
\begin{equation}
\phi _{\theta } \left ( b_{i} ,w_{j}  \right ) = b_{i}\times w_{j},
\end{equation}
where  \( b_i \) and  \( w_j \)represent individual visual and textual embeddings, normalized during training.

$L_{p}$ ensures a higher compatibility score between the grounding sentence and the true scene than between the sentence and any negative scenes (other point clouds in the mini-batch). The loss is formulated as:

\begin{equation}
\mathcal{L}_{p}(\theta)=
\mathbb{E}_{\mathcal{B}}\left[-\log \left(
\frac{e^{\phi_{\theta}(\mathbf{b}, w)}}
{e^{\phi_{\theta}(\mathbf{b}, w)}+\sum_{l=1}^{n} e^{\phi_{\theta}\left(b^{-}_{l}, \mathbf{w}\right)}}
\right)\right],
\label{ai:lvis}
\end{equation}
where \( b \) represents the visual embedding of the positive scene and $\left \{ b_{l}^{-}   \right \} _{l=1}^{n}$ are visual embeddings from the negative scenes. 

Similarly to $L_{p}$, $L_{t}$ encourages a higher compatibility score 
between the scene and the true grounding sentence compared to negative grounding sentences.
Negative grounding sentences are generated using a large language model. In our experiments, we adopt GPT-3~\cite{gpt}.


\subsubsection{Constructing Negative Grounding Sentences }
For a grounding sentence involving a target object \(s\) and its context \(c\), the goal is to replace \(s\) with an alternative object that fits the context \(c\) but inaccurately describes the actual scene. This ensures generating plausible yet incorrect grounding sentences.
For example, in the sentence ``A microwave is placed on the light wood-colored table,'' where \( s \) is ``microwave,'' we utilize a large language model to propose replacement objects.

The process consists of two primary steps:
Firstly, the language model generates the ten most plausible candidates for \( s \) based on the masked sentence template for \( c \).
Then, we manually remove candidates that either do not fit the scene or do not create a false grounding in the context. Therefore, we can generate negative grounding sentences such as ``An oven is placed on the light wood-colored table'' and remove negative grounding sentences like ``A fridge is placed on the light wood-colored table.'' By constructing these negative grounding sentences, we apply contrastive loss, which pushes the vision embedding away from the negative textual features. 

\textbf{Training with negative grounding sentences}
Using the generated context-preserving negative grounding sentences, we employ the negative contrastive loss $L_t$ as
\begin{equation}
\mathcal{L}_{t}(\theta)=\mathbb{E}_{\mathcal{B}}\left[-\log \left(\frac{e^{\phi_{\theta}(\mathbf{b}, w)}}{e^{\phi_{\theta}(\mathbf{b}, w)}+\sum_{l=1}^{n} e^{\phi_{\theta}\left(\mathbf{b}, w_{l}^{-}\right)}}\right)\right],
\label{ai:llang}
\end{equation}
where \( w \) represent the contextualized embedding of the true grounding sentence \( c\) and $\left \{ w_{l}^{-}   \right \} _{l=1}^{n}$ represent the embeddings of the corresponding negative grounding sentences $\left \{ c_{l}^{-}   \right \} _{l=1}^{n}$.


\subsection{Language-Guided Query Selection}
\label{sec34}
In DETr-like models, object candidate points play a crucial role in identifying the potential regions of the targets. 
However, previous works~\cite{eda, vpp} rely solely on the probability scores of the seed point features and often neglect the rich semantic information embedded in language queries.
To address this limitation, we design a language-guided query selection module that leverages language queries to generate object candidate points within the UR space. This is inspired by GroundDINO~\cite{groundinginfo}, a 2D vision-language model.
This module selects object candidate points that carry the same positional and semantic information as the input text.

Let $X_v \in R ^{N_v \times d }$ denote the visual queries and $X_t \in R ^{N_t \times d }$ denote the language queries. Here, $N_v$ is the number of visual queries,  $N_t$  indicates the number of language queries, and $d$ corresponds to the feature dimension. 
We aim to extract $N_q$ queries from visual queries to be used as inputs for the decoder. 
$N_q$ is set to be 256. The top $N_q$ query indices for the seed points denoted as $O$, are selected by
\begin{equation}
\mathbf{O}=\operatorname{Top}_{N_q}(\operatorname{Max}^{(-1)}(\mathbf{X}_{v}\mathbf{X}_{t}^{\top})),
\label{ai:lvis}
\end{equation}
where $\operatorname{Top}_{N_q}$ represents the operation to pick the top $N_q$ indices. 
$\operatorname{Max}^{(-1)}(\mathbf{X}_{v}\mathbf{X}_{t}^{\top})$ computes the maximum similarity between each visual query and all textual queries by taking the maximum along the last dimension of $\mathbf{X}_v \mathbf{X}_t^{\top} \in \mathbb{R}^{N_v \times N_t}$, where $N_v$ and $N_t$ are the numbers of visual and textual queries, and the symbol $^{\top}$ denotes matrix transposition, respectively.
The language-guided query selection module outputs $N_q$ indices. We can extract features based on the selected indices to initialize object candidate points. 

Similar to most object candidate points in DETR-like models\cite{eda}, the selected object candidate points \(O\) are fed into the cross-modal decoder to detect the desired queries and update accordingly. The decoded query \(Q\) is then passed through MLPs to predict the final target bounding box. 




\subsection{Training Objectives}
\label{sec35}
Following the previous work~\cite{eda}, the loss of Unispace-3D consists of the position loss $\mathcal{L}_{pos}$, the semantic loss for dense alignment $\mathcal{L}_{sem}$, the positive contrastive loss $\mathcal{L}_{pos}$ and the negative contrastive loss $\mathcal{L}_{neg}$, as:

\begin{equation}
\mathcal{L}= \mathcal{L}_{pos} + \mathcal{L}_{sem} + \gamma \left(\mathcal{L}_{pos} + \alpha \mathcal{L}_{p} + \beta \mathcal{L}_{t} \right).
\label{loss}
\end{equation}

The weights of each component in Eq.
\textcolor{red}{\ref{loss}} 
are discussed in Sec. 
\textcolor{red}{\ref{seclossweight}}.

\begin{table*}[h!]
  \begin{center}
    \caption{3D visual grounding results on the ScanRefer dataset. Accuracy is evaluated using IoU 0.25 and IoU 0.5. Methods marked with † indicate results reproduced using open-source code,
    while the others represent the best accuracies reported in their respective papers. Our single-stage implementation achieves higher accuracy without relying on an additional 3D object detection step (dotted arrows in Fig.~\textcolor{red}{\ref{fig:main}})}
    \label{tab: Scanrefer-main}
    \begin{tabular*}{\linewidth}{@{\extracolsep{\fill}} 
    c |c c|c c|c c} 
\toprule
       &    \multicolumn{2}{c|}{Unique($\sim 19\%$)} &  \multicolumn{2}{c|}{Multiple($\sim 81\%$)} &  \multicolumn{2}{c}{\textbf{Overall}}\\
      Method &0.25& 0.50& 0.25& 0.50&0.25& 0.50\\
\midrule
    Scanrefer\cite{scanrefer} & 67.64 & 46.19 & 32.06 & 21.26 & 38.97 & 26.10 \\
    ReferIt3D\cite{Referit3d}   & 53.8 & 37.5 & 21.0 & 12.8 & 26.4 & 16.9 \\
    TGNN\cite{TGNN}     & 68.61 &  56.80 & 29.84 & 23.18 & 37.37 & 29.70 \\
    InstanceRefer\cite{InstanceRefer}   & 77.45 & 66.83 & 31.27 & 24.77 & 40.23 & 32.93 \\
    FFL-3DOG\cite{FFL-3DOG} & 78.80 & 67.94 & 35.19 & 25.70 & 41.33 & 34.01 \\
    SAT\cite{sat} & 73.21 & 50.83 & 37.64 & 25.16 & 44.54 & 30.14 \\ 
    3DVG-Transformer\cite{3dvg} & 77.16 & 58.47 & 38.38 & 28.70 & 45.90 & 34.47\\
    3DJCG\cite{3DJCG} &  78.75 & 61.30 & 40.13 & 30.08 & 47.62 & 36.14\\
    MVT\cite{multiviewtrans}&  77.67 & 66.45 & 31.92 & 25.26 & 40.80 & 33.26\\
    BUTD-DETR\cite{bottom}   & 82.88 & 64.98 & 44.73 & 33.97 & 50.42 & 38.60\\
    3D-SPS\cite{3dsps}   & 84.12 & 66.72 & 40.32 & 29.82 & 48.82 & 36.98\\
    ViewRefer\cite{viewrefer} & -  &-  & 33.08 & 26.50 & 41.30 & 33.66 \\
    VPP-Net\cite{vpp} & 86.05 & 67.09 & 50.32 & 39.03 & 55.65 & 43.29 \\ 
 \midrule
    EDA\cite{eda}  †  & 84.07 & 67.23& 48.52 & 37.23 &53.82 &  41.71\\
    \textbf{UniSpace-3D}  & \textbf{86.72} &\textbf{ 70.07 }& \textbf{50.56 }& \textbf{39.89 }& \textbf{56.04} & \textbf{43.95 }\\ \hdashline
    EDA (single-stage) \cite{eda}  †  & 84.61 & 68.38& 46.65 & 36.91 &52.14 &  40.02\\
    \textbf{UniSpace-3D} (single-stage)   & \textbf{87.38} &\textbf{ 70.59 }& \textbf{49.60 }& \textbf{38.95 }& \textbf{55.34} & \textbf{43.17 }\\
    \bottomrule
    \end{tabular*} 
  \end{center}
\end{table*}

\section{Experiment} \label{sec:Experiments}
\subsection{Datasets}
We evaluate UniSpace-3D on the ScanRefer and ReferIt3D datasets. 
The \textbf{ScanRefer} dataset contains 51,583 descriptions of 11,046 objects across 800 ScanNet scenes. 
ScanRefer divides objects into ``Unique” and ``Multiple” subsets based on whether the object class is unique in the scenes.
The corresponding evaluation metric is Acc@IoU, which measures the fraction of descriptions where the predicted box and ground truth overlap with an IoU greater than 0.25 and 0.5.
The \textbf{ReferIt3D} dataset includes two subsets: Sr3D, which contains 83,572 template-generated expressions, and Nr3D, with 41,503 human-annotated descriptions spanning 707 scenes. Each scene in Sr3D/Nr3D can also be divided into ``Easy'' and ``Hard'' subsets depending on whether there are more than two instances.  
Following ReferIt3D\cite{Referit3d}, the primary evaluation metric for ReferIt3D is the accuracy of grounding predictions for textual descriptions. 

\begin{figure*}[htb]
  \centering
  \includegraphics[width=.95\linewidth]{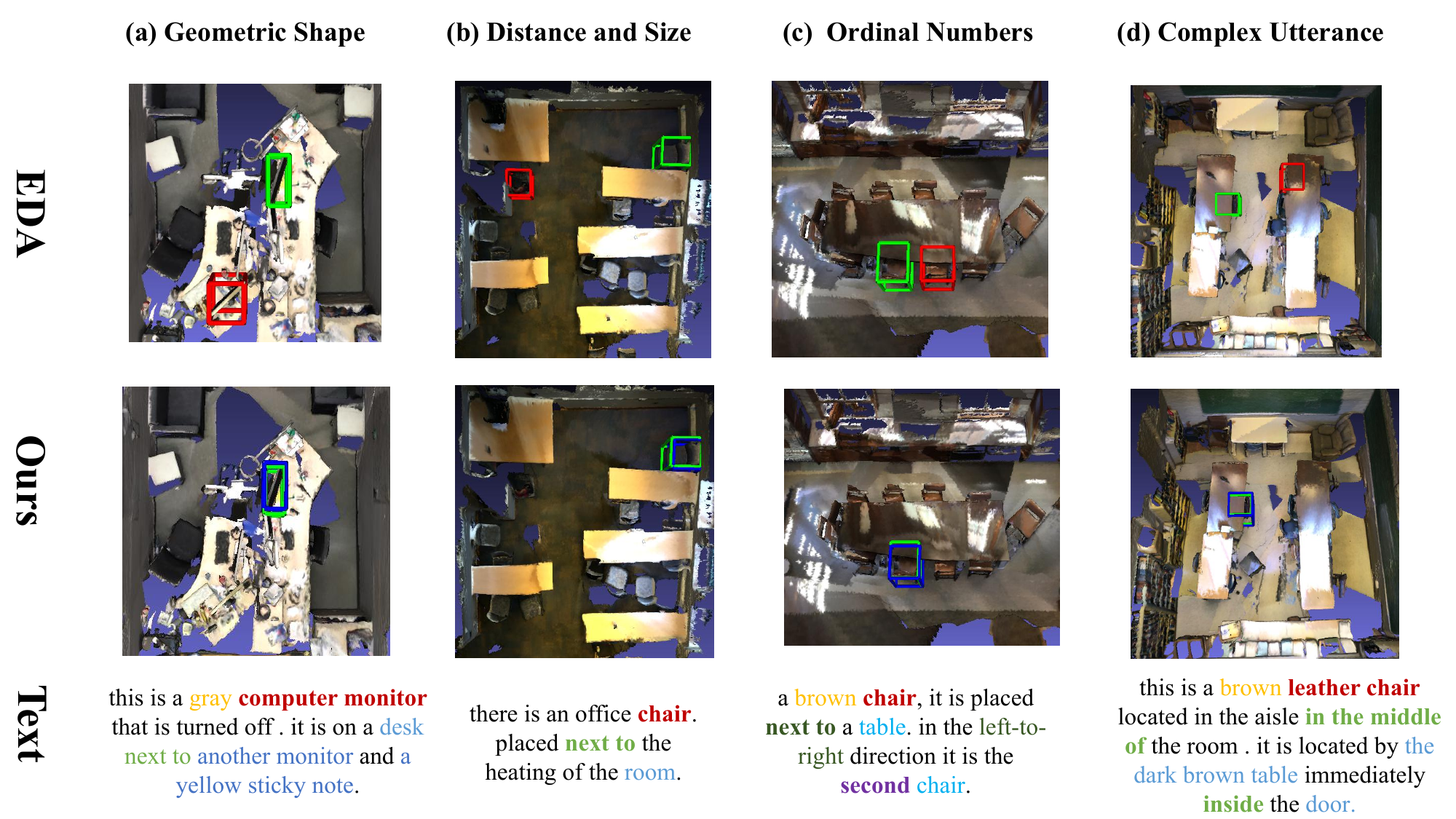}
  \caption{\label{fig:result}
        \textbf{Visualization of grounding results from different models on the ScanRefer dataset.}
         \textcolor{green}{Green} boxes represent ground-truth references. \textcolor{red}{Red} boxes show EDA results containing grounding errors (e.g., objects of the same category as the target). \textcolor{blue}{Blue} boxes represent proposals generated by our model.
          }
\end{figure*}

\subsection{Implementation Details}
For ScanRefer, the learning rate of the PointNet++  is $1e^{- 3} $. The learning rate of other modules is $1e^{- 4}$. It takes about 30 minutes per epoch, and around epoch 70, the best model appears.
The learning rates for SR3D are $3e^{- 4} $ and $3e^{- 5} $, with 50 minutes per epoch, requiring around 60 epochs of training.
The learning rates for Nr3D are $3e^{- 4} $ and $3e^{- 5} $, taking 30 minutes per epoch, and around 200 epochs are trained.
Since SR3D consists of concise, machine-generated sentences, it facilitates easier convergence. In contrast, both ScanRefer and NR3D are human-annotated, free-form, complex descriptions, which require more training time.
Codes are implemented by Pytorch and all experiments are conducted on two NVIDIA RTX GeForce 4090 GPUs.

\subsection{Quantitative Comparisons}
Tab.~\textcolor{red}{\ref{tab: Scanrefer-main}} presents the results of our experiments on the ScanRefer dataset, compared to previous works. UniSpace-3D outperforms all prior methods on both Acc@0.25IoU and Acc@0.5IoU, achieving $56.04\%$ and $43.95\%$, respectively, demonstrating a significant improvement. It surpasses our baseline EDA by $3.2\% $Acc@0.5IoU, and also $1.7\%$ higher than that of the VPP-Net\cite{vpp}. 

We report experimental results on the Nr3D and Sr3D datasets.
As shown in Tab.~\textcolor{red}{\ref{tab: ReferIt}}, our method achieves the highest accuracy of $57.8\%$ on Nr3D and $69.8\%$ on Sr3D, surpassing prior state-of-the-art methods.
In SR3D, since the language descriptions are concise and the object is easy to identify, our method achieves an accuracy of close to 70\%.
In Nr3D, descriptions exhibit noteworthy intricacy and detail, inducing additional challenges to the 3DVG task, our method still outperforms the EDA\cite{eda} by 5.1\%, thanks to the unified representation space for 3DVG.
Additionally, single-stage methods are excluded from the discussion, as ground truth boxes for candidate objects are provided in this setting.


\begin{table}[h!]
  \begin{center}
    \caption{Quantitative comparisons on the Nr3D and Sr3D datasets.}
    \label{tab: ReferIt}
    \begin{tabular*}{\linewidth}{@{\extracolsep{\fill}} 
    c |r r | r r} 
\toprule
        & \multicolumn{2}{c|}{Nr3D} &  \multicolumn{2}{c}{Sr3D}\\
       Method &Overall& Hard &Overall& Hard\\
\midrule
    TGNN& 37.3 &30.6 &45.0 &36.9 \\
    InstanceRefer &38.8 &31.8& 48.0& 40.5\\
        3DVG-Transformer & 40.8& 34.8 &51.4 &44.9\\
    SAT& 49.2& 42.4 &57.9 &50.0\\
    3D-SPS & 51.5& 45.1 &62.6 &65.4 \\
    MVT & 55.1& 49.1 &64.5& 58.8\\
    BUTD-DETR &54.6& 48.4& 67.0& 63.2\\
    EDA † & 51.9 & 45.8   &  67.1 &61.8\\
    VPP-Net & 56.9  & - &68.7  &-\\
    UniSpace-3D &\textbf{57.2} & \textbf{48.9} &\textbf{ 69.8}  & \textbf{63.7}\\
    \bottomrule
    \end{tabular*} 
  \end{center}
\end{table}

\subsection{Ablation Study}
\subsubsection{Ablation study on values of loss}
\label{seclossweight}
The representative results of a grid search over the weights in Eq.
\textcolor{red}{\ref{loss}}  
are summarized in Tab. 
\textcolor{red}{\ref{tab:weight}}.
Each line corresponds to a different weighting scheme for the components of the loss function. Notably, all configurations evaluated outperform the baseline method EDA~\cite{eda}, thereby validating the effectiveness and robustness of our proposed unified representation space.

As illustrated in lines (a) and (b), assigning equal weights to all components does not yield optimal performance.    This observation supports the notion that the different components contribute unequally to the overall objective and should thus be weighted accordingly. When giving $\alpha$ a higher weight (line (a)), it turns out that a weight that is too high would also lead to a decrease in performance, which may compromise the functionality of other components.

Through extensive tuning, we identify the weight configuration where $\alpha = 0.5$, while the other components are set to 0.3 and 0.1, respectively.     This configuration, denoted as option (c) in the table, achieves the best overall results and is therefore selected for use in our final implementation.

\begin{table}[htbp!]
  \begin{center}
    \caption{
    Grid search of the weight $\alpha, \beta$ and $\gamma$. Evaluated on the ScanRefer dataset. We select (c) for implementation.}
    \begin{tabular*}{\linewidth}{@{\extracolsep{\fill}} 
    c c c c | c c} 
    \toprule
      & $\alpha$&  $\beta$& $\gamma$  & @0.25IoU &@0.5IoU\\
    \midrule
      (a) & 1.0 & 1.0 &  1.0  &  53.95 &42.17 \\
      (b) & 0.5 & 0.5 &  0.5  &  54.91& 43.12 \\
      (c) & 0.5 & 0.3 &  0.1  & 56.04& 43.95\\
      (d) & 0.5 & 0.3 &  0.3  & 55.62 & 42.58\\
    \bottomrule
    \end{tabular*} 
    \label{tab:weight}
  \end{center} 
\end{table}

\subsubsection{Ablation study on introduced modules}
We use EDA as our baseline and conduct ablation studies to evaluate the effectiveness of each component in UniSpace-3D. Without further specification, all experiments are conducted on the ScanRefer validation set. 
The results of our experiments are presented in Tab. \textcolor{red}{\ref{tab: ablation}}.

\begin{table}[h!]
  \begin{center}
    \caption{Ablation study on different components of our model. `URE' denotes the unified representation encoder. `LGQS' refers to the language-guided query selection module. `MMCL' represents the multi-modal contrastive learning module.}
    \begin{tabular*}{\linewidth}{@{\extracolsep{\fill}} 
  c c c c |c c c} 
    \toprule
       &\multicolumn{3}{c|}{Different components } &  \multicolumn{3}{c}{Acc@0.5} \\
      & URE&  MMCL& LGQS &Unique &Multiple &Overall\\
    \midrule
   (a)   &   &   &     & 67.23& 37.23& 41.71\\
   (b)  &   \checkmark &   &   & 68.04  & 38.01  &42.39\\
   (c)  &   \checkmark & \checkmark   &   & 68.60  & 38.92  & 42.88\\
    (d)  &   &   & \checkmark  &69.04 & 38.62  & 43.17\\
   (e)   &   \checkmark &   &  \checkmark  & 69.51  & 39.81  & 43.65\\
   (f)   &   \checkmark &  \checkmark & \checkmark   & \textbf{70.07}  & \textbf{39.89}  & \textbf{43.95}\\
    \bottomrule
    \end{tabular*} 
    \label{tab: ablation}
  \end{center} 
\end{table}

For comparison, we train EDA~\cite{eda} based on the official publicly available code, and the results are in line (a).  The results demonstrate that URE improves performance by $0.81\%$ and $0.78\%$ in the “Unique” and “Multiple” splits. 
The improvements show that our unified representation encoder can effectively encode the relative
positional relationships and the relative semantic information.

We integrate the URE module into our baseline and individually modify or incrementally add each component to construct the experimental frameworks for testing. Experiments (c) and (f) add the multi-modal contrastive learning module (MMCL)  to further reduces the modality gap. Using MMCL boosts performance to 70.07\%, 39.89\%, and 43.95\%.  These results demonstrate the efficacy of multi-model contrastive learning in improving 3DVG performance.

Experiment (d) validates the efficiency of language-guided query selection module (LGQS). 
By generating object candidate points guided by language queries, LGQS emphasizes the key role of language queries in query generation. We precisely align the positions and semantics of target objects in both modalities, thereby facilitating a more accurate and reliable generation of object candidate points.

\subsection{Visualization}

\begin{figure*}[htb]
  \centering
  \includegraphics[width=.95\linewidth]{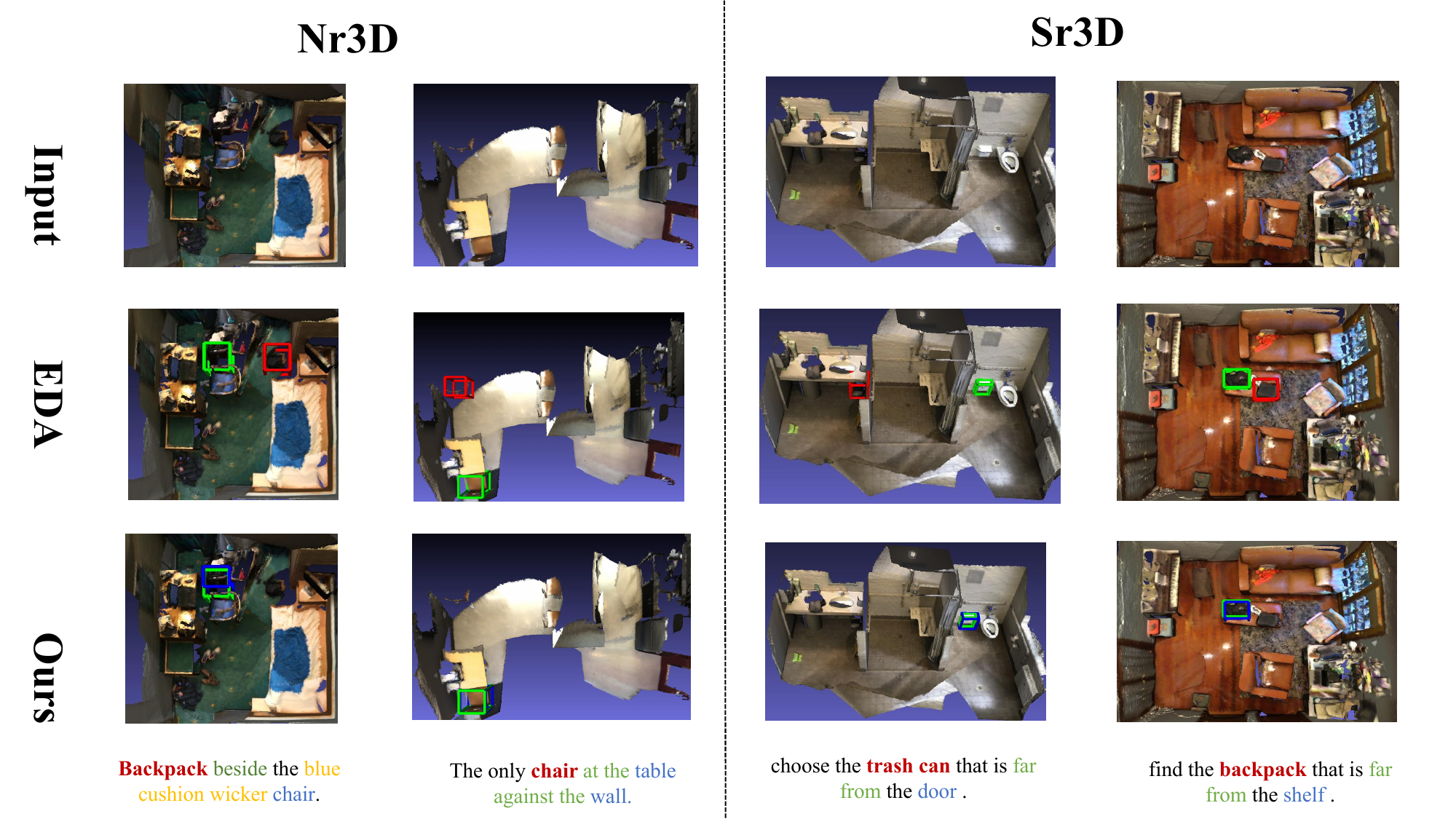}
  \caption{
    Qualitative comparison of the grounding results in the \textbf{Nr3d/Sr3D} dataset. For all boxes, \textcolor{green}{green} represents the ground-truth references; \textcolor{red}{red} represents EDA~\cite{eda} results containing grounding errors; \textcolor{blue}{blue} represents proposals generated by ours. Words in different colors show the results of text decoupling.
    }
\label{fig:nr3d and sr3d}
\end{figure*}

Fig.~\textcolor{red}{\ref{fig:result}} visualizes the results of four ScanRefer scenes, comparing predictions by EDA and UniSpace-3D to the ground truth. By comparing the visualization results, we clearly observed that Unispace3D effectively addressed four types of inaccurate positioning issues: geometric attributes, spatial distance or object size, ordinal numbers, and complex utterances. In each example, the green, red, and blue boxes represent the ground truth, EDA top-1 predictions, and our predictions, respectively. The results demonstrate the effectiveness of our method in understanding contextual information in the text to accurately identify the target objects. This improvement is made possible by the alignment of our textual embedding with visual embedding in the unified representation space.

\begin{figure}[htbp!]
  \centering
  \includegraphics[width=.98\linewidth]{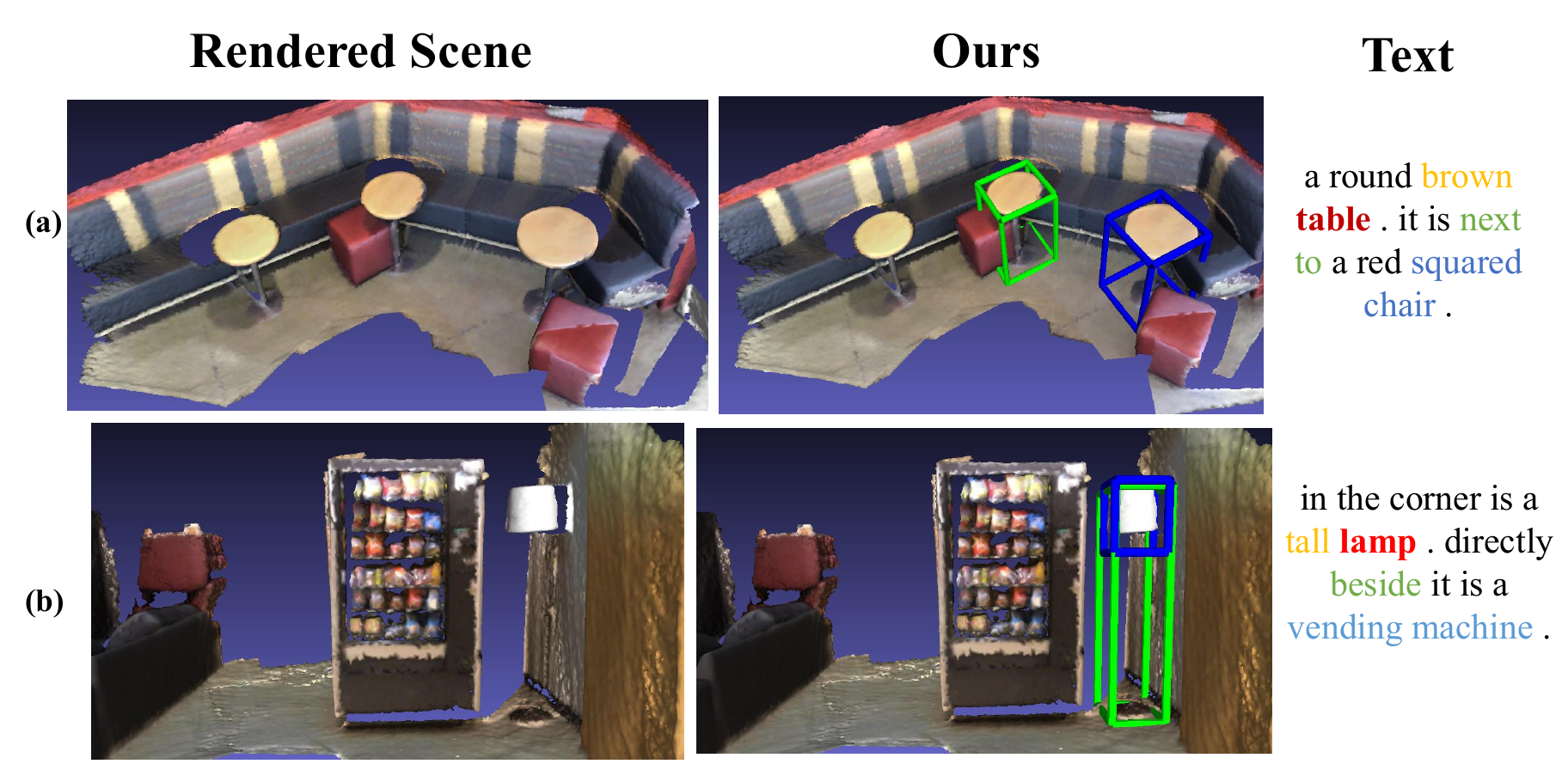}
  \caption{\label{fig:failure}
        \textbf{Qualitative results of some common failure cases.}
         \textcolor{green}{Green} boxes represent ground-truth references. 
         \textcolor{blue}{Blue} boxes represent proposals generated by ours.
          }
\end{figure}

The successful examples show that with the unified representation space for 3D visual grounding, the expression can better match the 3D scenes, resulting in more accurate groundings. 
This improvement is particularly evident in complex scenes with ambiguous or closely positioned objects, where our model demonstrates superior robustness and precision. 
More detailed qualitative results on Nr3D/Sr3D are detailed in Fig. \textcolor{red}{\ref{fig:nr3d and sr3d}}.
Qualitative results indicate that compared to EDA,
 our method exhibits a superior perception on the Nr3D/Sr3D dataset.
We also present two failure cases in Fig. \textcolor{red}{\ref{fig:failure}}. One occurs when text descriptions are ambiguous, and the other when point clouds are incomplete.

\section{Conclusion} \label{sec:Conculsion}
This paper introduces UniSpace-3D, a unified representation space for 3D visual grounding. UniSpace-3D leverages a pre-trained CLIP model to map visual and textual features into the unified representation space, addressing the inherent gap between the two modalities.  To enhance alignment, the multi-modal contrastive learning module minimizes the gap between visual and textual features.   Additionally, the language-guided query selection module identifies object candidate points matching natural language descriptions.  Extensive experiments demonstrate that UniSpace-3D improves performance by at least 2.24\% over baseline models. These demonstrate its effectiveness in bridging vision and language for 3D visual grounding , which also highlights its potential as a foundation for future research in multi-modal 3D understanding and embodied AI research.


\bibliographystyle{IEEEtran}
\bibliography{arxiv}

\vfill

\end{document}